\documentclass[runningheads,a4paper]{llncs}

\usepackage{amssymb}
\usepackage{graphicx}

\usepackage{url}

\usepackage{multirow}
\usepackage{graphicx}
\usepackage{caption,subcaption}
\usepackage{amsmath}
\usepackage{amsfonts}
\usepackage{multirow}
\usepackage{slashbox}

\newcommand{\vt}[1]{
\mathbf{#1}
}

\renewcommand{\eqref}[1]{
  Eq. \ref{#1}
}

\urldef{\mailsa}\path|{son.tran,qing.zhang,mohan.karunanithi}@csiro.au|
\newcommand{\keywords}[1]{\par\addvspace\baselineskip
\noindent\keywordname\enspace\ignorespaces#1}

\begin{document}

\mainmatter  

\title{On Multi-resident Activity Recognition in Ambient Smart-Homes}

\titlerunning{On Multi-resident Activity Recognition in Ambient Smart-Homes}

%
%
\author{Son N. Tran%
 \and Qing Zhang \and Mohan Karunanithi}
\authorrunning{Tran et al.}

\institute{The Australian E-Health Research Centre, CSIRO, Brisbane, QLD 4026, Australia\\
\mailsa\\
}

%
%

\toctitle{Lecture Notes in Computer Science}
\tocauthor{Authors' Instructions}

\maketitle

\begin{abstract}
Increasing attention to the research on activity monitoring in smart
homes has motivated the employment of ambient intelligence to reduce
the deployment cost and solve the privacy issue. Several approaches
have been proposed for multi-resident activity recognition, however,
there still lacks a comprehensive benchmark for future research and
practical selection of models. In this paper we study different
methods for multi-resident activity recognition and evaluate them on
same sets of data. The experimental results show that recurrent neural
network with gated recurrent units is better than other models and
also considerably efficient, and that using combined activities as
single labels is more effective than represent them as separate
labels.

\end{abstract}

\section{Introduction}
In intelligent environments such as smart homes activity recognition
plays an important role, especially when applying to health monitoring
and assistance \cite{Das_2004}. Many efforts has been made to model
the activities of residents in order to facilitate reasoning of their
behaviour. The success of such models would result in reducing cost of
traditional health care, a smarter and safer home for eldercare, and
better assistance for patients. Classifying human activities has been
studied intensively within computer vision domain
\cite{Poppe_2010}. This, however, may raise an issue on privacy of
residents due to the use of unwelcome devices,
i.e. cameras. Alternatively, many other approaches rely on wearable
sensors \cite{Plotz_2011}, which seems less intrusive but require
users to wear an electronic device everywhere and everytime.  Recent
attention is aiming at intelligent environments where residents can
live their own way, without being disturbed by the presence of a
device on their bodies. This is an important research topic that would
shape the future of smart homes. With the advance in pervasive sensing
technologies one can install a set of non-intrusive sensors in the
environment with respect to residents' privacy
\cite{Wilson_2005,Singla_2010}. However, in contrast to the
development of ambient hardware, the reality of intelligent algorithms
for such modern smart homes is still challenging.

Activity recognition in ambient environment has been studied for
years, most of that focuses on single resident, aiming to support
independent living \cite{vanKasteren_2008}.  However, in practice this
is not always the case since modern smart environments should be able
to support multiple occupants. As a result, there is a growing desire
for a model that is capable of capturing the complexity nature of both
independent and joint activities. This is a challenging task because
different from the case of single resident where the sensors' states
reflect directly the activity of a specific person, in multi-resident
case that information, as known as {\it data association} is
not commonly known. In recent work, temporal approaches have been
widely employed to model activities in smart homes (see the survey
\cite{Benmansour_2015}). However,
there still lacks a comprehensive study on how different sequence
models perform in this application domain.

In this paper we investigate the use of hidden Markov models (HMMs),
conditional random fields (CRFs), and recurrent neural networks (RNNs)
for multi-resident activity recognition in ambient smart homes. The
study also focuses on two different methods of encoding activities of
multiple residents: combined labels and separate labels.  We expect
that the work would serve as a benchmark and guideline for those who
seek for a solution in this domain. In general, we model the activity
sequences of multiple residents as an input of a function with the
input is the states of a smart home. Such states in this case are
defined as the values of the sensors in the smart home. The output can
be divided into different types of representation. First, we consider
the output as a single variable which is the combined activities of
all residents. With this we can apply sequence models such as HMMs,
CRFs, and RNNs for classification task straightforward. The second
type of output is encoded by separating it into multiple variables,
with each presenting the activities of a resident, while sharing the
same input. In order to deal with multiple residents we need more
complex models to capture the interactive and collaborative
behaviours. For HMMs, we employ the factorial variant
\cite{Ghahramani_1997} add more cross dependencies. In the case CRFs,
cross dependencies would be computationally expensive especially for
the data of very long sequences, so that factorial CRFs
\cite{Sutton_2007} are used. For RNNs, we share the hidden layer for
all different label variables. In the experiments, we conduct
evaluations of the models on three smart homes from two benchmark
datasets. The results show that recurrent neural network with gated
recurrent units is better than other models and also considerably
efficient. We also found that using combined labels is more effective
than separate-labels.

The paper is organised as follows. In the next section we review the
related literature of our work. Section \ref{sec:mrar} describes the
multi-activity modelling framework and the models studied in this
papers. In section \ref{sec:exp} we perform experiments and analyse the
results. Finally, in Section \ref{sec:concl} we conclude the work.

\section{Related Work}
Research on multi-resident activity recognition has been emerging
recently due to the increasing demand for health monitoring in ambient
intelligent environments. The task can be done by employing sequence
models to perform prediction on the activity events over time. Hidden
Markov Models \cite{Rabiner_1990} is a popular statistical model for
sequential data.  It is characterised by the dependency of an
observation variable on a hidden variable at each time step, and the
dependency of the hidden variable itself on its previous state. HMMs
can be employed for activity recognition easily. In particular, one
can define the observation as the sensors state, i.e. video frame,
wearable or/and ambient sensors' values, and the hidden variable as
the activity \cite{Kim_2010}. In multi-resident smart homes, HMMs have
been studied intensively, as being showed in previous works
\cite{Alemdar_2013,Chen_2014,Singla_2010,Cook_2012}. A straightforward
approach is to use a single HMM for combined activities, i.e. treating
the activities of all residents as a random variable. For example, the
activities can be combined as joint labels so that they can be
represented by a single hidden variable \cite{Chen_2014}. Another
method to model the activities of multiple residents is to create
multiple HMMs, one for each resident \cite{Chiang_2010}. Such model,
as known as parallel HMM, has been evaluated in the case that data
association is provided. This means that the observation has been
separated for each resident and only represents the sensors which are
associating to that resident. The disadvantage of this model is the
hidden variables of all HMMs are independent from each others. In
multi-resident environments, however, there always exist correlation
and interaction between the residents. This issue is addressed by
adding the crossed dependencies to the hidden variables in all
HMMs. By coupling such HMMs one can assume that the activity of a
resident is dependent not only on his previous activity but also on
the previous activities of other residents.  There was a proposal of
coupled HMM and factorial HMM in computer vision domain
\cite{Brand_1997}, but only coupled HMM was employed for sensor data
\cite{Chiang_2010}. Besides HMMs, CRFs \cite{Crandall_2008,Hsu_2010}
and incremental decision trees (IDT) \cite{Prossegger_2014} also have
been used for multi-resident activity recognition .

From learning perspective, the problem of multi-resident activity
recognition can be seen as multi-tasks learning on sequence
data. However, most of the work we found in literature focus on
modelling different tasks from different data sourses by taking the
advantage of recurrent neural networks in learning more generalised
representation from larger amount of data combined. Different from
that, in this work we do not have such augmentation since there exist
only one dataset for activities of multiple residents.
\section{Multi-Resident Activity Modelling}
\label{sec:mrar}
Let us denote $a^{m,t}$ and $o^t$ as the activity of resident $m$ and
the sensors' state at time $t$ respectively. For ease of presentation
we denote $\vt{a}^t=\{a^{1,t}, a^{2,t}, .. , a^{M,t}\}$ as the
activities of all $M$ residents at time $t$. We use $t_1:t_2$ to
denote a sequence of events/states from time $t_1$ to $t_2$. For
example, $\vt{a}^{t_1:t_2}=\{\vt{a}^{t_1}, .. , \vt{a}^{t_2}\}$ is the
sequence of activities performed by all residents from time $t_1$ to
$t_2$. In this paper, we evaluate two ways of modelling the activities of
multiple residents. First, we combine the activities such that the
activities of all resident at a time step is represented by a single
variable. For that we need to predict $\vt{a}^{1:T}$ given the states
of sensors $\vt{o}^{1:T}$. Second, we model  each resident's
activity as a separate variable.

\subsection{HMM-based Approaches}
\begin{figure}[ht]
  \begin{subfigure}{0.32\textwidth}
    \includegraphics[width=1\textwidth]{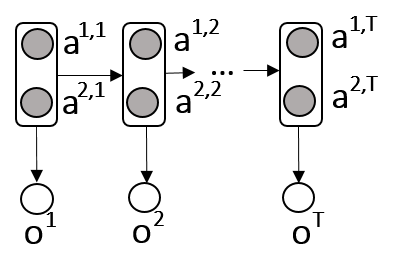}
    \vskip -.4cm
    \caption{hmm}
    \label{hmm}
  \end{subfigure}
  \begin{subfigure}{0.32\textwidth}
    \includegraphics[width=1\textwidth]{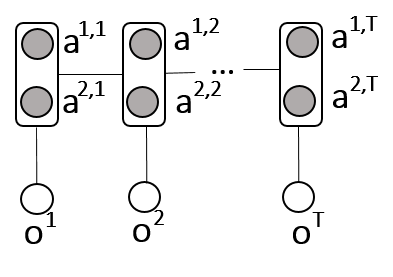}
    \vskip -.4cm
    \caption{crf}
    \label{crf}
  \end{subfigure}
  \begin{subfigure}{0.32\textwidth}
    \includegraphics[width=1\textwidth]{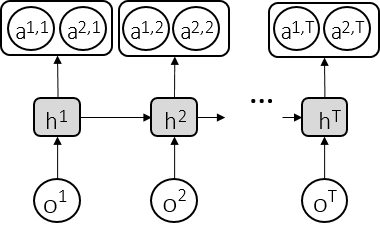}
    \vskip -.3cm
    \caption{rnn}
    \label{rnn}
  \end{subfigure}
  \\
  \begin{subfigure}{0.32\textwidth}
    \includegraphics[width=1\textwidth]{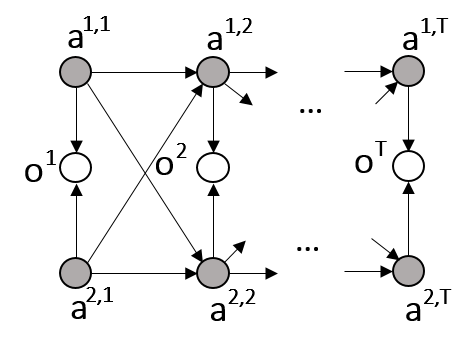}
    \vskip -.4cm
    \caption{fhmm}
    \label{fhmm}
  \end{subfigure}
  \begin{subfigure}{0.32\textwidth}
    \includegraphics[width=1\textwidth]{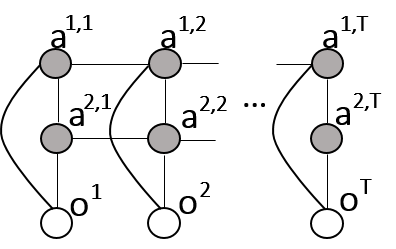}
    \vskip -.3cm
    \caption{fcrf}
    \label{fcrf}
\end{subfigure}
  \begin{subfigure}{0.32\textwidth}
    \includegraphics[width=1\textwidth]{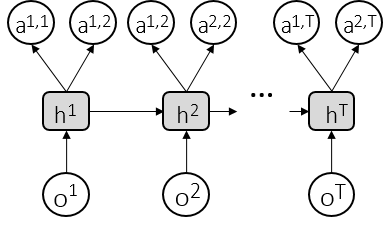}
    \vskip -.3cm
    \caption{mrnn}
    \label{mrnn}
  \end{subfigure}
  \vskip -.3cm
  \caption{Sequence models for multi-resident activity recognition. hmm: hidden Markov model; crf: conditional random field; rnn: recurrent neural networks; fhmm: factorial hidden markov model (with cross dependencies); fcrf: factorial conditional random field; mrnn: multi-labels recurrent neural networks. The top row depicts the models for combined labels and the bottom row depicts the models for separate labels.}
  \vskip -1cm
\end{figure}

A HMM \cite{Rabiner_1990} consists of a single hidden  and an
observation variable  which assumes a Markov process. 
In the case of combined labels we can use a single HMM to model the activities as a joint distribution as:
\vskip -.3cm
\begin{equation}
  p(\vt{a}^{1:T},o^{1:T}) = p(o^1|\vt{a}^{1})p(\vt{a}^{1})\prod_{t=2}^Tp(o^t|\vt{a}^{t})p(\vt{a}^{t}|\vt{a}^{t-1})
\end{equation}
 \vskip -.3cm Inference of activities given a sequence of sensors'
 states can be done efficiently using dynamic programming,
 i.e. Viterbi algorithm \cite{Rabiner_1990}. For the separate labels,
 different HMMs have been used such as parallel HMMs, coupled HMMs
 \cite{Wang_2011,Son_2017}. In this paper we use factorial HMM with
 cross dependency shown in Figure \ref{fhmm}, as this variant achieves
 better performance than the other HMMs \cite{Son_2017}. Factorial HMM
 \cite{Ghahramani_1997}, is a HMM with multiple hidden variables. In
 order to represent the relations between activities among residents,
 we add cross connections from all hidden variables at time $t-1$ to
 each hidden variable at time $t$. This results in a factorial HMM
 model with cross dependency as we introduce here in the paper. The
 joint distribution of this HMM is:
\vskip -.3cm
\begin{equation}
  p(\vt{a}^{1:T},o^{1:T}) = p(o^1|\vt{a}^{1})\prod_m p(a^{m,1})\prod_{t=2}^T(p(o^t|\vt{a}^{t})\prod_mp(a^{m,t}|\vt{a}^{t-1})) 
\end{equation}
\vskip -.3cm
Similar to a normal HMM, inference of activities can easily done by dynamic programming. Here, only the transition and the prior probabilities are changed in
comparison to the HMM above. Therefore, we
can apply the Viterbi algorithm  by replacing
$p(\vt{a}^t|\vt{a}^{t-1})$ with $\prod_m p(a^{m,t}|\vt{a}^{t-1})$ and $p(\vt{a}^1)$ with
$\prod_m p(a^{m,1})$.
\label{sec:impl}
\subsection{CRF-based Approaches}
Conditional random field is a probabilistic graphical model which can be used for modelling sequences, similar as HMMs. The difference here is that a CRF is a discriminative model representing a conditional distribution:
\begin{equation}
    p(\vt{a}^{1:T}|o^{1:T}) = \frac{1}{Z(o^{1:T})}\prod_{t}\Psi(\vt{a}^t,\vt{a}^{t-1},o^t)
\end{equation}
where $\Psi(\vt{a}^t,\vt{a}^{t-1},o^t) = \exp(\sum_k \theta_k
f_k(o^t,\vt{a}^t,\vt{a}^{t-1}))$ with $f_k$ are the feature functions,
$\theta_k$ are parameters of the features, and $Z(o^{1:T})$ is the
partition function:
\begin{equation}
Z(o^{1:T}) = \sum_{{\vt{a}'}^{1:T}} \prod_{t}\Psi(o^t,{\vt{a}'}^t,{\vt{a}'}^{t-1})
\end{equation}
This is a CRF for combined labels, for the case of separate labels, we
split the hidden unit to have a variant as known as factorial CRF
\cite{Sutton_2007}. In Figure \ref{fcrf} we illustrate the graphical
presentation of this model.
\subsection{RNN-based Approaches}
A recurrent neural network is constructed by rolling a feed-forward
neural network over time where the hidden layer is connected to itself
by a recurrent weights. As shown in Figure \ref{rnn}, we can use the
output layer to represent the combine activities of multiple
residents. For example, at time $t$ the probability of a joint
activity $\vt{a}^{t}$ is:
\begin{equation}
  p(\vt{a}^{t}|o^{1:t}) = \text{softmax}(\vt{h}^{t}U + \vt{b})
\end{equation}
where $U$ is the weight matrix connecting the hidden layer and the
output layer; $\vt{b}$ is the biases of the output units. We will show
how hidden state $\vt{h}^{t}$ is computed later in this section. For
the other case where activities of residents are modelled separately
we can split the output layer into multiple layers, as shown in Figure
\ref{mrnn}. Let us suppose that there are $M$ residents, the
probability of a resident $m$ performs an activity $a^{m,t}$ at time t
is: $p(a^{m,t}) = \text{softmax}(\vt{h}^{t}U_m + \vt{b}_m)$. Here each
output layer is connected with a shared hidden layer by a weight
matrix $U_m$. The hidden state in both cases (combined labels and
separate labels) is computed as $\vt{h}^{t} =\tanh(o^{t}W +
\vt{h}^{t-1}V+ c)$. This is the simplest form of hidden unit which is
said not very useful to capture long-term information and suffer the
problem of vanishing/exploding gradient \cite{Hochreiter_1997}. This
is also shown than such problems can be ameliorated by using complex
gates in the hidden units. In this work, we will empirically
investigate three type of hidden units for RNNs, including the
original ``tanh'' and two others with more complex gates as known as
long-short term memory (LSTM) \cite{Hochreiter_1997} and gated
recurrent units (GRU) \cite{Cho_2014}.
\section{Experiments}
\label{sec:exp}
In the experiments we evaluate the effectiveness of all the models above on three smart homes data in two benchmark datasets. 
The CASAS data\footnote{\url{http://ailab.eecs.wsu.edu/casas/}} was
 collected in the WSU smart department Testbed with two residents
 where each resident performing 15 unique activities
 \cite{Cook_2010}.   The data is collected in 26 days in a smart home
 equipped with 37 ambient sensors.
 The ARAS data\footnote{\url{http://www.cmpe.boun.edu.tr/aras/}}
\cite{Alemdar_2013} is collected in two different houses, denoted as
House A and House B, in 30 days. In these environments, there are 20
sensors for two residents in each house where each resident is asked
to perform 27 different activities.
\subsection{Evaluation}
We denote $\hat{\vt{a}}^{1:T}$ with $\hat{a}^{i}=\{\hat{a}^{1,t},
\hat{a}^{2,t}, ..., \hat{a}^{M,t}\}$ as the predicted activities of
all residents in the house for each instance in the test set
$\mathcal{D}_{test}$. We also denote a ground truth is
$\vt{a}^{1:T}$. The performance of a model is measured by the accuracy
of each resident's activities and the accuracy of all residents'
activities. The former is computed as:
\begin{equation}
  accuracy_m = \frac{1}{|D_{test}|}\sum_{a^{m,1:T} \in \mathcal{D}_{test}}\frac{1}{T} \sum_t(a^{m,t}==\hat{a}^{m,t})
\end{equation}
where $a^{m,t}$ and $\hat{a}^{m,t}$ are ground truth and predicted activity of resident $m$ at time $t$ in each instance of the test set $\mathcal{D}_{test}$. Similarly, the accuracy for activities of all residents is:
\begin{equation}
  accuracy_{all} = \frac{1}{|D_{test}|}\sum_{\vt{a}^{1:T} \in \mathcal{D}_{test}}\frac{1}{T} \sum_t(\vt{a}^{t}==\hat{\vt{a}}^{t})
\end{equation}
\subsection{Results}
\label{sec:results}
We partition the CASAS data into $24$ days for training, $1$ day for validation and $1$ day for testing. The ARASA and ARASB are the data from ARAS House A and ARAS House B each consist of $7$ days for training, $2$ days for validation and $2$ days for testing.
\begin{table}[ht]
 \vskip -.5cm
 \begin{center}
   {\footnotesize
  \begin{tabular}{|l||c|c|c||c|c|c||c|c|c||c|}
    \hline
    \hline
        {\backslashbox{Model}{Data}}& \multicolumn{3}{|c||}{CASAS}& \multicolumn{3}{|c||}{ARAS House-A}& \multicolumn{3}{|c||}{ARAS House-B} & Average\\
    \hline
    &R1 & R2 & All & R1 & R2 & All & R1 & R2 & All& \\
    \hline
    \hline
    RNN$_{tanh}$  & 66.62  & 64.88  & 58.08 & 67.02  & 73.09  & 53.07 & 81.09  & 78.16  & 76.15            & 68.68\\
    \hline
    mRNN$_{tanh}$ & 69.14  & 60.61  & 47.94 & 68.12  & 74.74  & 53.26 & 91.93  & 78.99  & 76.86            & 69.06\\
    \hline
    RNN$_{gru}$  & 92.26  & 87.68  & 83.66 & 69.79  & 74.20  & 56.23 & 81.69  & 78.95  & 76.83            & \textbf{77.92}\\
    \hline
    mRNN$_{gru}$ & 90.89  & 83.97  & 77.91 & 70.03  & 75.37  & 56.65 & 82.04  & 78.90  & 76.83            & 76.95\\
    \hline
    RNN$_{lstm}$ & 89.40  & 87.45  & 82.14 & 69.94  & 73.59  & 56.44 & 82.15  & 77.84  & 76.72            & 77.29\\
    \hline
    mRNN$_{lstm}$& 69.05  & 84.30  & 77.49 & 69.97  & 75.39  & 56.25 & 81.66  & 78.92  & 76.60            & 74.40\\
    \hline
    HMM  & 65.24  & 65.82  & 56.58 & 43.95  & 54.93  & 19.13 & 79.29  & 76.98  & 75.07            & 59.67\\
    \hline
    fHMM & 73.55  & 67.44  & 55.43 & 44.19  & 54.58 & 19.13 & 79.17   & 77.10 & 75.07             & 60.63\\
    \hline
    CRF  & 76.40  & 66.07  & 64.32 & 70.73  & 78.17  & 61.72 & 88.36  & 89.27  & 76.23            & 74.58\\
    \hline
    fCRF & 58.21  & 56.76  & 45.84 & 69.50  & 69.50  & 55.95 & 76.01  & 76.01  & 74.44            & 64.69\\
    \hline
    \hline
  \end{tabular}
  }
  \end{center}
  \vskip -.3cm
  \caption{Prediction accuracy for all models on three datasets. R1, R2, All are accuracy of predicted activities of resident 1, resident 2 and the joint activities}
  \label{tab:all_results}
  \vskip -1cm
\end{table}

The models are selected as follows. For HMM and fHMM we selected the best models based on the Laplacian smoothing factor. The smoothing factor is chosen from $10^{-6}$ to $10^{-2}$ in log-space. For the CRFs, we do not use any hyper-parameters and set the maximum iteration is $1000$. We use MALLET to implement fCRFs and set the penalty hyper-parameters to zeros. For the recurrent neural networks, we perform model selection by using grid-like search on number of hidden units within $\{10,50,100,500,1000\}$, and learning rate from $0.0001$ to $1$ in log-space. If the optima is not apparent we expand the search. The RNNs are trained using stochastic gradient descent with early stopping. Due to the need for initialisation of the RNN models we repeat each experiment on recurrent neural networks $50$ times and report the average results. We denote RNN and mRNN are recurrent neural networks for combined labels and separate labels respectively. We also use subcripts $tanh$, $gru$, $lstm$ to denote the type of hidden units in RNNs.

Table \ref{tab:all_results} shows the results of all models on three
datasets. In CASAS data, RNN$_{gru}$ outperforms other models. In
ARASA, HMM based models have very low performance. It seems that the
simplicity of HMM cannot capture the complexity of this data as we
learn that the number of observed sensors' values in ARASA is 10 times
more than CASAS and 3 times more than ARASB. In this dataset, CRF has
the highest accuracy. In ARASB most of the models have similar
performance with fCRF achieves the lowest accuracy of $74.44\%$ and
mRNN$_{tanh}$ achieves the highest accuracy of $76.86\%$. We observe
that RNNs with complex gates (GRU and LSTM) seem to overfit the
training and validation sets since they have much more paprameters
than RNN with tanh units. In the ``Average'' column are the mean
accuracy of individual activities and the joint activities in all
datasets. We can see that among all models, RNN$_{gru}$ achieves the
best performance overall. This is because, the small size of the
experimental data makes the compactness of RNN$_{gru}$ advantagous
over RNN$_{lstm}$ which has larger mber of parameters.
\begin{figure}[ht]
  \centering
  \includegraphics[width=1\textwidth]{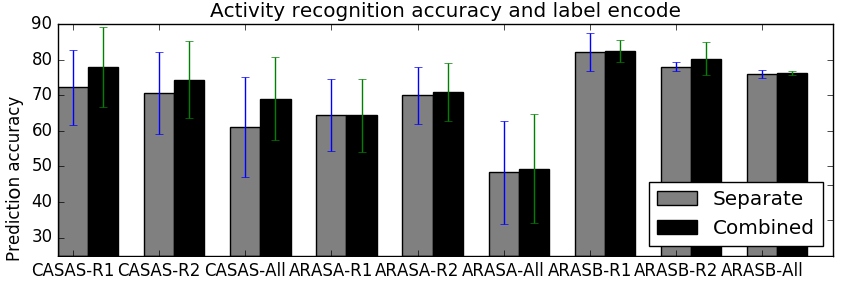}
    \caption{Combined labels v.s separate labels for multi-resident activity recognition}
    \label{fig:combined_vs_separate}
\end{figure}
\subsection{Combined labels versus Separate labels}
We now analyse the effectiveness of combined labels v.s. separate labels. We compute the average accuracy of all models that use combined labels approach on each data and compare it with the separate labels approach. The results are demonstrated in Figure \ref{fig:combined_vs_separate}. It is consistent that models with combined labels have better prediction accuracy than models with separate labels. An interesting finding from the results here is that the combined label approaches not only have higher  accuracy for joint activities but also outperform the separate label approach in predicting individual activities.

\subsection{Efficiency}
Finally, we analyse the efficiency of the models in this application. Theoretically, the combined labels may need more parameters than the separate labels as the former require variables to represent $K^1\times K^2 ... K^M$ activities while the latter use $M$ variables each represent $K^m$ activity values. However, this might be different in practice where smaller combined label model may have better results than bigger separate-label models. It would depends on how models are selected by validation sets. In Table \ref{tab:time} we report the running time of the models which give the best results in section \ref{sec:results}. The most effective model is HMM which only needs less than one minute to complete the training and prediction for any of the three datasets. fHMM is slightly less efficient than HMM. However, it should be noted that both HMM and fHMM are the least effective models in Table \ref{tab:all_results}. fCRF is the slowest model among the others. Although the Java implementation of MALLET might be the reason of this inefficiency  the prediction performance of fCRF is only better than HMM and fHMM. So the use of this model in practice should be questioned, especially when its results are even lower than CRF.
\begin{table}[h]
  \begin{center}
  \begin{tabular}{|c|c|c|c|}
    \hline
    \hline
        {\backslashbox{Model}{Data}}& {CASAS}&{ARAS House-A}& {ARAS House-B}\\
    \hline
    \hline
    \hline
    RNN  & 96.23  sec  & 3064.50 sec  & 3547.74 sec \\
    \hline
    mRNN & 406.31 sec  & 3.39 hrs     & 3.44 hrs \\
    \hline
    GRU  & 203.58 sec  & 1.31 hrs     & 1.43 hrs \\
    \hline
    mGRU & 545.39 sec  & 1.47 hrs     & 1.38 hrs \\
    \hline
    LSTM & 183.64  sec & 7.87 hrs     & 8.61 hrs \\
    \hline
    mLSTM& 638.35  sec & 6.54 hrs     & 1.71 hrs \\
    \hline
    HMM  & 0.17 sec    & 95.23 sec    & 50.74 sec \\
    \hline
    fHMM & 0.23 sec    & 97.52 sec    & 50.77 sec \\
    \hline
    CRF  & 64.32 sec   &  $\sim$8 hrs & 289.17 sec \\
    \hline
    fCRF & 3.6 hrs     & $\sim$419 hrs& $\sim$220 hrs \\
    \hline
    \hline
  \end{tabular}
  \end{center}
  \caption{(Average) Computational time for best models in section \ref{sec:results} to train and predict activities in three datasets. We denote the time in hours (hrs) if it is more than 3600 seconds otherwise we denote it in seconds (sec).}
  \label{tab:time}
\end{table}
Overall, among all models, RNN$_{gru}$ would be the best choice since it has the best performance while being considerably efficient. It is not as fast as RNN$_{tanh}$ but it is quicker than the other RNN based models in most cases. Especially, it outperform HMM based models and fCRF models with large margin. CRF is also a good choice since its best models  are even faster than RNN$_{gru}$ in CASAS and ARASB. However, in these two datasets it has lower accuracy than RNN$_{gru}$ while in ARASA it is much slower than RNN$_{gru}$.
\section{Conclusions}
We have presented a benchmark study on activity recognition for multi-resident smart homes with ambient sensors. We empirically show that recurrent neural network with gated recurrent units is better than other models and also considerably efficient. We also show that using combined activities as single labels is more effective than represent them as separate labels.
\label{sec:concl}
\bibliographystyle{plain}
\bibliography{../../bibs/bibio}
\end{document}